\documentclass[letterpaper]{article} 

\usepackage{notaaai2026}  

\newcommand{\appdx}[1]{(see Appendix #1)}

\usepackage{times}  
\usepackage{helvet}  
\usepackage{courier}  
\usepackage[hyphens]{url}  
\usepackage{graphicx} 
\usepackage{natbib}  
\usepackage{caption} 
\usepackage{algorithm}
\usepackage{algorithmic}

\usepackage{newfloat}
\usepackage{listings}
\DeclareCaptionStyle{ruled}{labelfont=normalfont,labelsep=colon,strut=off} 
\floatstyle{ruled}
\newfloat{listing}{tb}{lst}{}
\floatname{listing}{Listing}
\usepackage{amsmath}
\usepackage{framed}
\usepackage{fancyvrb}
\usepackage{graphicx}

\newtheorem{expml}{Example}

\title{Why Isn't Relational Learning Taking Over the World?}
\author{David Poole}
\affiliations{Dept of Computer Science, University of British
  Columbia\\
  Vancouver, BC, Canada\\
  \url{https://www.cs.ubc.ca/~poole/}
  }
\begin{document}
\maketitle
\begin{abstract}
  
  Artificial intelligence seems to be taking over the world with systems that model pixels, words, and phonemes.  The world is arguably made up, not of pixels, words, and phonemes but of entities (objects, things, including events) with properties and relations among them. Surely we should model these, not the perception or description of them. You might suspect that concentrating on modeling words and pixels is because all of the (valuable) data in the world is in terms of text and images. If you look into almost any company you will find their most valuable data is in spreadsheets, databases and other relational formats. These are not the form that are studied in introductory machine learning, but are full of product numbers, student numbers, transaction numbers and other identifiers that can't be interpreted naively as numbers. The field that studies this sort of data has various names including relational learning, statistical relational AI, and many others. This paper explains why relational learning is not taking over the world -- except in a few cases with restricted relations -- and what needs to be done to bring it to it's rightful prominence.
  
\end{abstract}

\section{Introduction}
AI has hit the news recently with models that
learn to predict images, text, sound and video. Many have even speculated that
such technologies will lead to artificial general intelligence (AGI),
and be more intelligent than people.

In my AI courses, I often include the clicker question of Figure
  \ref{clickerfig}. Reading machine learning books and papers would lead one to assume the answer is
  A. Reading the press on generative AI would lead one to think
  the answer is B. Very
  few of the students choose A or B.
  
 \begin{figure}
\begin{framed}
  What is the real world made of?
  \begin{itemize}
  \item[A] Features or random variables
    \item[B]
      Words, pixels, phonemes \dots
    \item[C] Entities and events (e.g., plants, people, diseases,
      lectures, university courses)
    \item[D] Huh? There is a real world?
    \end{itemize}
  \end{framed}
  \caption{A clicker question}
  \label{clickerfig}
\end{figure}

  Learning and reasoning about entities, events
  and relations among them is central to AI:
\begin{quotation}\em
The mind is a neural computer, fitted by natural selection with
combinatorial algorithms for causal and probabilistic reasoning about
plants, animals, objects, and people.
\begin{flushright}
-- Steven \citet[][]{Pinker:1997ua} 
\end{flushright}
\end{quotation}


By relational learning, I mean learning models that make probabilistic
predictions about entities (things, objects, including events), their properties and
relations among them. The aim is to model entities rather than modelling
their manifestations in 
language or images.


Part of the theme of this paper is that we should not lose connection
to the downstream task that
a learned model is used for. Learning is usually not an end in
itself, but is used for decision making. This could be for
autonomous decision making, where probabilities and utilities are needed, or
human-supported decision making where multiple scenarios can be evaluated by a human decision maker.

\section{Relational Data}
If you were to examine company data, environmental data, or government
data -- with few exceptions -- you will find that the most valuable
data not in natural language text or images but in
spreadsheets, relational databases and other relational formats. This
relational data is typically not the sort of tabular data used in introductory machine
learning courses -- for which standard  algorithms such as gradient tree boosting \citep{Chen:2016aa}
work directly on -- but is full of product numbers, student numbers,
  transaction numbers and other identifiers. These identifiers are often
  represented as numbers, but the numbers themselves don't convey 
  meaning. For gradient tree boosting to be effective for such data,
  it needs to exploit 
  relational structure \citep{Natarajan:2012aa}.

The simplest form of relational data is exemplified by collaborative
filtering datasets such as used in the Netflix challenge
\citep{Bell:2007nf} and Movielens datasets
\citep{Harper:2015}. Movielens contains \emph{rating} datasets containing user-movie-rating-timestamp
triples, where the user and movie are identifiers represented by
arbitrary numbers. There are also tables for user properties and
movie properties, including links into the Internet Movie Database (IMDB). An even
simpler version is a \emph{rated} dataset that ignores the actual ratings,
and is just a user-movie dataset (perhaps with timestamps). To predict
rating (or rated) from just the rating dataset, you can use
latent properties (embeddings) of users and of movies to predict ratings. Representing latent properties as a
vector, matrix
factorization \citep{Koren:2009aa} learns about movies and users, without learning general knowledge. Another possible technique is to do a self-join, which can
result in a user-user table or a movie-movie table.

Most real-world databases are much more complicated, with multiple relations.

When considering relational data, the following questions are often implicit and
affect what is represented:
\begin{itemize}
\item Does the system know what exists? Most databases assume that
  they know what exists; things that exist are given 
  identifiers, and what doesn't exist is ignored. In many real-world
  cases, we don't know what exists. Was there a person in the house
  last night?  If so, who? Who threw
  the rock that broke the window? (It might have not been a rock
  or maybe no one threw it.)
\item Do different identifiers denote different entities? This is known
  as the \emph{unique names assumption} and features in logic
  programming and most databases. Identifying equality (two
  descriptions describing the same entity) arises in
  record linkage \citep{reason:FelSun69a} and citation matching
  \citep{reason:Pasula03a}. It is a big problem in medicine where drug dispensers
  need to decide whether someone asking for drugs is the same person
  who asked previously.
\end{itemize}


\section{Knowledge Graphs}
(Subject, verb, object) triples are the basis for knowledge graphs.
The subject is an entity. The verb
is a relation or property. The object is an entity or a datatype value, such as
a number
or a string. A collection of triples can be treated as a labelled directed graph where the entities
and datatype values are nodes. There is an
arc from the subject to the object, labelled with the verb.
Triples are universal representations of relations: any relation can
be represented as subject-verb-object triples 
\citep[See e.g.,][Section 16.1]{Poole:2023aa}.

\begin{figure}
 
  \begin{center}
\includegraphics[width=\columnwidth]{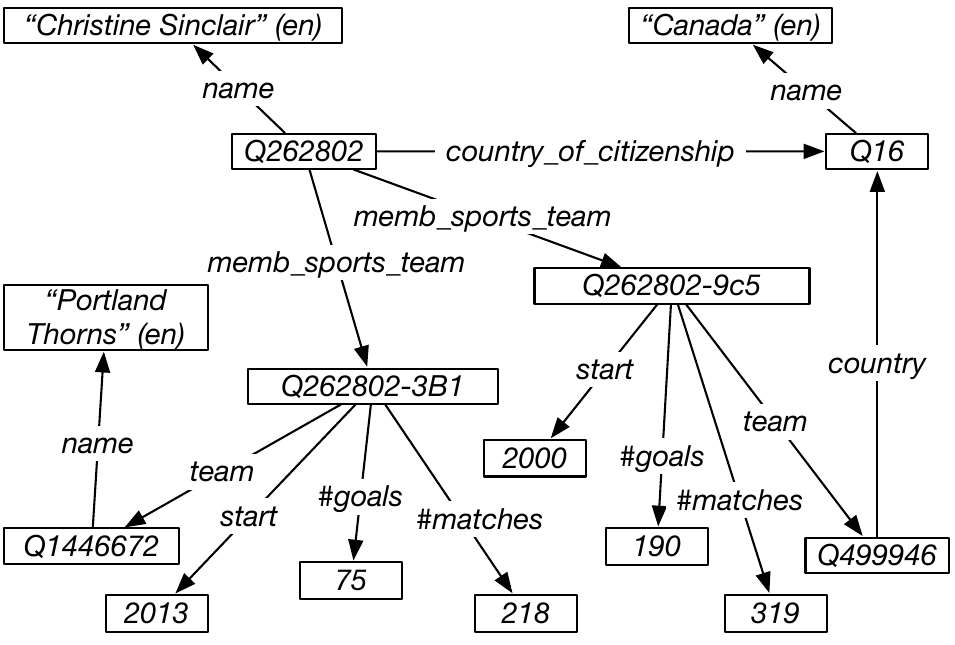}
\end{center}

\caption{ Part of the
    Wikidata knowledge graph about Christine Sinclair; from Figure 16.1 of \citet{Poole:2023aa} }
  \label{wikidata-fig}
\end{figure}

One large public knowledge graph is Wikidata \citep{Wikidata:2014aa}.
Table \ref{wikidata-tab} shows tables about the
football (soccer) player Christine Sinclair. She is the player (man or
woman) with the most goals in international play. Figure
\ref{wikidata-fig} shows part of the corresponding Wikidata knowledge graph. In Wikidata, Christine Sinclair is
represented by the identifier Q262802. The country Canada is
represented by the identifier Q16.

 \begin{table}
\begin{center}
  \begin{tabular}{lll}\hline
    entity & name & language \\\hline
    Q262802 & ``Christine Sinclair'' & en\\
    Q16 & ``Canada'' & en\\
    Q1446672 & ``Portland Ferns'' & en
  \end{tabular}
\end{center}
     \begin{center}
  \begin{tabular}{lllll} \hline
member & team & start & \#matches &
                  \#goals\\\hline
    Q262802 & Q499946 & 2000 & 190 & 319\\
    Q262802 & Q1446672 & 2013 & 218 & 75
  \end{tabular}
\end{center}
\caption{Facts about football player
  Christine Sinclair (Q262802) in tabular form}
\label{wikidata-tab}
\end{table}

This shows two ways of converting relations to triples. The first is
to group columns such as ``Canada''-en, giving
(Q16, name, ''Canada''-en).
Grouping columns does not work in
general as there are exponentially
many grouping of columns. It is only
done in restricted cases.

The second way to convert relations to triples is to use (row, column, value) triples. The row can be a
primary key if there is one, or is a new identifier -- a
\emph{reified} entity. Reify means ``make into
an entity'' (or ``thingify''). In Figure
\ref{wikidata-fig}, Q262802-3B1 and Q262802-9c5 are
reified entities. 
The number of facts about a reified entity is the number of columns in the
table, which is typically just a handful.
To convert to the original relation, one can do a database join of the
triples on the
reified entities and project onto the other properties (which is how I
generated the tables from the Wikidata triples). Reification does not
lose information. It makes the representation more modular: other
properties, including meta-properties (such as the author) can be
easily added to a knowledge graph.

Another way to convert relations into triples is to pair  
attributes, called the
star-to-clique method
\citep{Wen:2016aa}. This, however, loses information.
Figure \ref{fb15k-qs-fig} shows some examples of the sort of triples
that arise from this method.

Note that graph learning \citep{Hamilton:2020aa} and relational learning are isomorphic problems.
A graph can be represented as a  tail-head-label \emph{arc} relation.
Relations can be represented as knowledge graphs. These are complementary fields.
It is not clear that a typical relational database is anything like a typical graph.
Knowledge graphs tend to be much more heterogenous than graph benchmarks \cite{Bechler-Speicher:2025aa}; the only general structure is that induced by reification.
Many of the questions of interest are different. For example, there tends to be only a single knowledge graph so graph matching \cite{Roy:2025aa} is  not really  a problem of interest.
Most of the issues in this paper do not apply to graph learning in general.

\section{Datasets}
One of the most used knowledge graphs in the literature is  FB15k \citep{Bordes:2013aa},
a large subset of Freebase. Freebase \citep{bollacker2008freebase} was a free and open knowledge base developed
from 2007-2016. It was acquired by Google, who donated the contents to
Wikidata, and use another branch as the Google knowledge
base. In building FB15k, \citet{Bordes:2013aa} selected entities and relations that appeared in more
than 100 triples. They also created a larger knowledge graph, FB1M,
with the most frequently occurring 1 million entities, which has been
used much less in subsequent research.

\begin{figure*}
\begin{Verbatim}
(A.S. Livorno Calcio, /soccer/football_team/current_roster, 
         Forward (association football))
(Forward (association football), /soccer/football_roster_position/team, 
         Cambridge United F.C.)
(California, /location/religion_percentage/religion, Methodism)
(Ambient music, /music/genre/artists, Portishead (band))
(Marriage, /people/marriage/spouse, Noel Gallagher)
(Hannah Montana: The Movie,
      /film/film_regional_release_date/film_release_region, Egypt)
\end{Verbatim}
\caption{Some of the triples in FB15k. The subject and object
  entities have their identifier replaced with the name. The verb
  (property) names
  have been abbreviated.}
\label{fb15k-qs-fig}
\end{figure*}

Figure \ref{fb15k-qs-fig} shows some of the triples in 
FB15k. The last two are obviously projections of relations that are
more interesting, telling us who Noel Gallagher was married to (and
probably when) and when Hannah Montana: The Movie was released in
Egypt.

We\footnote{Bahare Fatemi, Mehran Kazemi, Ali Mehr and Martin
  Wang. See \url{https://cs.ubc.ca/~poole/res/FB15k-triples.txt} for
  example triples for all 1344 predicates.}
created English
version of the test set that,
together with the plausible examples that existing methods ranked highly, could be used to crowdsource 
ground truth. The ground truth could then be used to evaluate prediction
methods. However, we decided
determining whether a triple was true wasn't a skill people could do
quickly and accurately, even
when they had access to Wikipedia, Google and other online
sources. Positive examples were straightforward
, but determining something
is false requires guessing.

\citet{toutanova-chen-2015-observed}  
noticed that a predictor could do reasonably well on FB15k  by just determining that
two predicates were (approximately) equivalent or inverses. To determine whether a
predictor was doing more than this, they created FB15k-237 with the
inverse functions removed (it would remove one of the first two triples of
Figure \ref{fb15k-qs-fig}). FB15k-237 isn't better than FB15k; they
test different things. Together they provide interesting evidence: the
difference tells us whether a predictor can recognize
inverses; the performance on FB15k-237 tells us whether the method
can exploit other regularities. We would want a predictor that can determine
inverses, but would like it to do more than that.

Including only most frequent entities means that there are \emph{no}
reified entities in the resulting triples. In Wikidata, over 98\% of the
entities appear as the subject of fewer than 10 triples 
  \appdx{A}. Training and evaluating on datasets
where all such entities are removed will not lead to better predictors
on more realistic  datasets.

Big data often implies more entities, but less data, on average, about each
entity. As entities are added, they start off as stubs with few
triples, resulting in a long tail of entities about which little is known. 
As more
detail is added about an entity, there are more associated reified entities
with inherently few triples
\appdx{A}.

Another standard dataset is WN18
\cite{Bordes:2013aa}, based on WordNet \cite{Miller:1995aa}, a large
lexical database of English. The entities are words, and the relations
include hyponym (subclass), hypernym (superclass), part\_of and type, with
loosely connected hierarchies for nouns, verbs, adjectives and adverbs.  Most of the
relations are transitive. I would not expect that words are
prototypical entities.

Graph learning
also has issues with standard benchmarks \cite{Bechler-Speicher:2025aa,Roy:2025aa}.  

JF17K \citep{Wen:2016aa},
FB-AUTO, M-FB15K \citep{Fatemi:2020aa} and RelBench \citep{Robinson:2024aa}  are benchmarks that do
not convert to triples.
Such datasets are a start in building more realistic test sets.

\section{Relational Predictions}
As far as relational learning is concerned, it is important to distinguish two types of
relational datasets.

In the \emph{complete knowledge} or \emph{closed
  world} assumption,  statements  not in a database are false. E.g., Airlines know the actual departure and
arrival times of flights
in the past, who is
    booked on a flight,  and who showed up for boarding. They do not
    know who will show up tomorrow.
When data is considered
complete up to some time,
the predictive problem is to
predict the future from the past.
Consider, for example, the problem of
predicting the passenger load or whether I will miss a connection, of a trip from Vancouver BC to
Boston MA on July 2 2026. This is a tricky prediction as July 2 is a
special day --
is one day after a holiday in Canada, and 2 days before one in USA --
and 2026 is unusual in that many Canadians are boycotting USA and the
FIFA World Cup is being held then; Vancouver and
Boston are host cities. Rather than guessing, we should
be able to make an informed prediction based on data.

Surprisingly, there is very little work on making predictions about
the future from arbitrary complete databases. We can't add an
airline reservation system to a learner and get it to predict future demand. It might be because most company
databases are proprietary and include much metadata, such as the
database schema. Individual companies such as Amazon and Netflix have
huge datasets they are mining, but the techniques are  proprietary
and are undoubtedly tuned to the particular task, such as maximizing
profit and
engagement, rather than allowing general queries of what is predicted.
\citet{Steck_Baltrunas_Elahi_Liang_Raimond_Basilico_2021} describe
techniques used by Netflix. Feature engineering (such as carefully chosen
joins and projections) can created
tables that methods such as XGBoost or deep learning can work on.

The alternative to the complete knowledge assumption is the \emph{open-world assumption}: 
the data is not complete and there is missing data. Wikipedia contains a very small
proportion of the true statements even about those entities chosen to be
notable enough to be included. Building predictive models from such models has two main challenges:
\begin{itemize}
\item There are no negative statements.
\item Missing data is not missing at random.  
\end{itemize}

Databases do not store arbitrary relations. Some relations are stored
and others are derived. Which relations to store is carefully designed
\citep{Codd:1970aa}.
Database designers
choose to store only true statements for some of the possible relations, selected to make the database small. They don't represent
``did-not-buy'' or ``has-watched-movie-starring''  as explicit
relations, instead choose smaller
relations to explicitly represent, and use queries and rules to infer
other ones. 

The lack of negative statements is a feature of databases, but with
the open world assumption, it makes
predicting probabilities problematic. There is no way to estimate probabilities without external
information about the entities and relations not represented (e.g., knowing that people are typically married to zero
or one person, or that celebrities are richer than average). This meta information cannot be assumed
to be explicit in a
relational database.


      Predicting property values, when the set of values is a constant
      set, such as the positions of football players, may be amenable to techniques
      for tabular learning, such as gradient tree boosting, following
      feature engineering. However, when making predictions where the
      values are entities (called \emph{link prediction}) this is infeasible. 
        

\section{Training}
A standard way to make probabilistic predictions is to use a softmax or sigmoid output and to optimize log loss (an observation  with predicted probability $p$  incurs a
cost of $\log p$, the expectation of which is cross entropy) on the
training set with some regularization.

In relational datasets with no
negative statements, the prediction that optimizes log loss is to predict
everything is \emph{true}, with a loss of zero on any data set that only
includes positive statements. However for a test set that includes
statements that should be false, 
this prediction has an infinite test loss.
In training, the lack of negative statements is generally handled by
a form of contrastive learning: adding random triples and using these as negative examples (the few
that might be correct are noise). The number of negative versus
positive examples provides
input that is just fiction. Unlike a prior probability, this input
is not overwhelmed by observed data.

Relational models differ from standard tabular models in two main ways:
\begin{itemize}
  \item parameter sharing or weight tying; namely \emph{logical variables}
and \emph{quantification} in
logic approaches
\citep{reason:Poole93g,De-Raedt:2007aa,Domingos:2009},
\emph{exchangeability}\footnote{Identifiers can be exchanged and
  the probability distribution should not change. Exchangeability of
  identifiers does not imply exchangeability of random variables \citep{Finetti:1974aa}.} \cite{niepertB14}, or
the \emph{convolutional} in convolutional
graph neural networks \citep{Schlichtkrull:2017ac}. 
\item aggregation, where a property of an entity or relation among
  entities depends on other entities, such as
  predicting the
  gender or age of a person from the movies they have watched.
\end{itemize}


When considering relational data as a graph, the neighbours of an entity are typically either all entities or those known to be related by existing relations.
When considering entities $n$ hops away from an entity, the number of entities is exponential in $n$. This is, of course, bounded by the number of entities.
Having all entities as neighbours of an entity allows for richer hypotheses about which pairs are related, but impacts on what methods are appropriate. 
With large knowledge bases, algorithms that are quadratic in the number of entities are impractical. Learning needs to be done in linear time, at most, and inference in sublinear time. Fortunately many questions can be approximated by sampling.

There are a myriad of, often ingenious, learning techniques used for
relational data which I won't go into.

\section{Evaluation}

The evaluation of a knowledge graphs depends on what is asked. To
predict whether a random triple is in Wikidata, answering ``no'' is
over 99.99999999995\% accurate\footnote{Wikidata has
  about 1.65 billion triples on 117 million
  items and 290,000 properties
  (\url{https://grafana.wikimedia.org/d/000000175/wikidata-datamodel-statements}). With
  no domain or range constraints in Wikidata, any entity can be
  the subject or object of any property. (Let's ignore strings and other
  datatypes as values, as there are too many possible values). There are thus $(1.17e8)^2*2.9e5 \approx 4e21$ possible
  triples, of which $1.6e9/ 4e21 = 4e{-}13$ are in Wikidata. 
}. Accuracy of
random triples is not a reasonable measure. Most random
triples are nonsense, which no one would ask and could be answered just
by knowing the types of the entities. 

One could evaluate predictions using log loss, as in the training outlined
above. Those papers that report log loss
typically do the following: Given a test triple, mutate it by replacing the
subject or object to create a negative example. The predictor is
then evaluated on these positive and negative examples. (The 
examples generated to be negative that are actually true are just
noise.)  Unfortunately the probabilities are just fiction, reflecting
the statistics of the knowledge base, not the statistics of the world.

The most common way to evaluate a knowledge graph predictor is to use
ranking: given  test triple
$(s,v,o)$, ask the predictor to rank
the entities related to a given subject $s$ and verb $v$, i.e., to
predict which triples of the form $(s,v,?)$ are true
(and similarly to rank
the subjects for a given verb and object). Given a total ordering of predictions
$o_1,o_2,o_3, \dots$, the rank is the $i$ such that $o=o_i$.
In filtered prediction \citep{Bordes:2013aa}, the $o_i$ that are
known to be true
(because the corresponding triple is in the training or test set) are removed
from the ordering.

The mean rank is one possible measure but
is dominated by a few very large ranks, and is not used much any more.
A common measure is \emph{hit-at-$k$} for some integer
$k$ (such as 1, 3 or 10), the proportion of test cases where the rank of the correct
answer is $k$ or less. 
The $k$ in hit-at-$k$ seems like an arbitrary parameter. A method
that is common and seems to not have arbitrary parameters is the mean
reciprocal rank (MRR) \cite{Radev:2002aa}. This weights the $k$th rank
as $1/k$. It can be interpreted as an agent who always considers the
first item, considers the second item half the time, the 3rd item 1/3
of the time on so on. However this interpretation is problematic \appdx{B}.

Ranking has many problems, including:
\begin{itemize}
\item Some questions cannot be asked, such as ``who is the Pope married
  to?'' (where the answer is ``nobody''). 
\item  Asking $(s,v,?)$ leaks information about the test set. Asking who $e$ is married to tells us that $e$ is a
  person, an adult and not the Pope. Knowing that someone rated a
  movie, as can be obtained from asking for a prediction for the rating of a movie, is valuable information that can be exploited to produce
  better predictions. 
  One could imagine a task that provides information about an entity, as a prompt to an LLM
  might, and asks questions about that entity.  For many
  tasks, questions should not provide information. Whether leaking
  information about the test set is appropriate depends on the
  downstream application.

\item Some queries are trivial; for example, with fewer
  than 10
  football (soccer) positions in FB15k and most teams having all of them,
  asking for a position in a team (see Figure
  \ref{fb15k-qs-fig}), any list of all positions will give a
  hit-at-10.
  \item Some queries are almost impossible, for example asking which
  soccer team has someone playing \emph{forward} -- one of the queries of
  the FB15k test set -- means listing all soccer teams. Guessing which
  one the query is about is almost impossible. You would expect that
  an omniscient agent -- one who knows everything -- would
  go well in any reasonable evaluation. However, an omniscient agent who knows all
  soccer teams and all players and their positions, would not be able
  to guess what team the query was about.
\item The actual probabilities are lost. One predictor might be very
  sure of one answer, and have to fill in the rest. Another predictor
  might have a handful of plausible answers, they are not sure
  about. These predictors could produce the same ordering. There
  is no penalty for being overconfident or reward for being sure (and correct) or for actually specifying
  uncertainty, even though this is valuable information for making decisions. 
\item It loses sight of the downstream task. 
  Considering top-$k$ is reasonable for cases where the results
can be checked, such as in movie recommendation, web search or
suggesting to a doctor the $k$ most
likely diagnoses so they can be reminded of something they may have overlooked. It seems less useful when just presenting $k$ possible
answers for which there is no easy way to check which, if any, are correct.
State-of-the-art methods that combines existing models
\citep{pmlr-v235-li24ah,10737292} have hit-at-10 rates on FB15k-237
of 55.8\%. and 52.58\% respectively, which does not seem
very useful for any real-world task where correctness cannot be
easily verified.
\end{itemize}




\section{Looking Forward}

There are many success stories from instances of relational
learning from protein prediction \cite{Jumper:2021aa}, where the
atoms are entities and relations includes \emph{bonds}, to predicting
congestion on roads and at intersections for route-planning
\citep{Delling:2015aa} to the combination of ML and the internet of things in
agriculture \citep{SHARMA2024100292,Mureithi:2024aa}. These build
models particular to the application. Generalizing particular models
so we can make predictions from arbitrary relational datasets
and queries is
still in its infancy, despite years of research.  This provides a
great opportunity for researchers to make fundamental contributions.
For relational learning to reach its full potential, the following (at
least) need to be done:

We need to consider real public datasets on problems that someone cares
about. Unfortunately someone caring means that the datasets are often
valuable  and so propriety. Relational learning is
ideal for making predictions on electronic health records
\cite{Weiss:2012aa,Natarajan:2013aa}, but such records are confidential and cannot be made
public for other researchers.
Many of the current public datasets are public
because no one (currently) cares about them. One class of datasets that should be exploited more is
environmental datasets produced and published by government agencies,
such as European Environmental Agency\footnote{See e.g.,
  \url{https://www.eea.europa.eu/en/datahub}}. These often contain all
measurements taken before a certain time, even if they are not
complete with respect to related causes of these. These datasets, however, are often
overwhelming for researchers who want to concentrate on the general
problem of relational learning, and are much more difficult to use than
the datasets in the ML repositories \citep[but see][]{tschalzev2025unreflectedusetabulardata}.
What is useful to actually predict from these is not obvious for non-domain
experts. We need to interact with diverse sets of domain experts to
find useful tasks.  

The main use of predictions is make decisions. One of the greatest
inventions of the 20th century was utility theory \cite{reason:VonMor53a}: utilities are
measures of preference that can be combined with probabilities. To
make decisions, we thus
need probabilistic predictions and utility functions. For relational
models, we need better
evaluation to handle the predicament that we want to predict
probabilities but only have positive statements. One solution
is to include and exploit meta-information, such ``there are no
more triples involving this subject and  verb''. There is a continuum between closed-world and open-world assumptions; 
datasets are often complete for some narrow part of the world.

For incomplete knowledge bases, we need to explicitly consider why data is
missing. Having test examples that are a
random subset of all of the triples, does not provide a surrogate
to real-world problems where data is invariably not missing at
random  \citep{rubin1976inference}. I doubt there are any Taylor Swift
officially released
albums missing from Wikidata, but for most of the recording artists on Wikidata
most of their recordings are missing. Probabilistic models of missing
data \citep{Marlin:2011,Mohan:2013aa} tend to be heavyweight,
requiring detailed modelling of why data is missing. This is difficult when
virtually all of the triples are missing.
We made an attempt to build a lightweight model of missing
data \cite{Poole:2020vv} with the aim of extended it to relational
models. 

When predicting entities, rather than assuming that the correct answer
is one of the known entities, there are three possible types of
answers: one or more of
the known entities, an entity or entities not represented, or that there
is no entity. Consider asking for birth mother. For most
people in a knowledge base the answer is ``someone not
represented''. Asking for a child of a person, a person may have (perhaps multiple)
children represented, 
children not represented (or both) or have no children. Similarly, asking for
an event that caused something, it might be a known event, an unknown
event or there might no event. An unknown event may need to be given
an identifier to give it properties. A prediction should not be
judged on the actual identifier assigned.

We need to be clear about whether we want to predict general knowledge
or learn about the entities in the training set. Unless the test sets
are carefully designed, there will tend to be a bias towards
learn about just one of these. Many models such as probabilistic logic
programs \citep{reason:Poole93g,De-Raedt:2007aa} and convolutional
graph neural networks \citep{Schlichtkrull:2017ac} learn what looks
like general
knowledge that is then be applied to the training set to predict
properties and relations of the entities in training set. Tests sets
from the same dataset do not test whether these models
generalize to new populations or related domains.

To have statistics,
we need to combine the data in at least one dimension. E.g., we could
ask for the average congestion at a particular intersection at a
particular time, or the average congestion at intersections on a
particular date and time, but not at at a particular intersection at a
particular date and time, as there is at most a single data point. There could be latent features (embeddings) for intersections
and dates, but some intersections and dates are peculiar and it might be
better to learn about particular entities and not just general knowledge.




Many methods rely on embeddings/latent features, often fixed-sized
embeddings. It seems bizarre to have the same size embeddings for the
USA and the relationship between Sinclair and the Portland Ferns
(Figure \ref{wikidata-fig}). Somehow we need the embedding/latent complexity
to depend on the complexity of (the information about) the entity.

Aggregation (e.g., predicting the gender of a user from the movies they watched)
is the Achilles heel of relational learning. Most
models either have a built-in aggregation such as noisy-or for
probabilistic logic programs \citep{reason:Poole93g} or logistic
regression for weighted logical formula models
\citep{Domingos:2009,Kazemi:2014aa}, use a distribution of related entities in
relational attention \citep{velickovic2018graph,Wang:2019aa} or use explicit operations such a
sum or mean \citep{Schlichtkrull:2017ac}.  They are problematic
theoretically (e.g., 
asymptoticly) \cite{Poole2014a,Buchman:aaai2015,Adam-Day:2024aa} and in
practice \cite{Kazemi:2017ub}. They need to work with no related
entities and as the number of related entities goes to infinity. The
models either implicitly assume the related entities provide
independent evidence (as do sum, noisy-or and logistic regression),
or act the same as if there were one or few related entities (as do max, average and attention).
Determining whether the evidence is independent (so can be
accumulated) or is dependent (in which case extra evidence can be
discounted) is difficult, and invariably ignored.

To realize the full potential of relational learning, we need to
combine the information from \emph{all} datasets, and build predictive models
from these. Combining multiple heterogenous datasets is the challenge
of the semantic web \cite{Berners-Lee:2001aa}. Building predictive
models from heterogenous data might seem impossible, but we know how
to; it's called
\emph{science}.  To combine all recorded experimental and
observational data to make better predictions,  we need at least
ontologies\footnote{Scientific ontologies include Open Biological
  and Biomedical Ontology \url{http://obofoundry.org}, WHO Family of
  International Classifications 
  \url{https://www.who.int/standards/classifications}, SMOMED-CT
  healthcare ontology \url{https://www.snomed.org}} (to allow semantic interoperability)
and provenance information \citep{gil-etal-ess16,Sikos:2021aa}, including
what was manipulated, what was controlled for, what was observed and
when, how was what to record decided (modelling the missing
information), who manipulated the data, etc. Science involves building hypotheses,
typically from multiple datasets,
that persist, ready to be revised or rejected as new evidence comes
in. Hypotheses can be about anything; from the effects of global warming,
to what foods are edible, to the traffic in Vancouver. There needs to be multiple hypotheses for a dataset,
including the null hypothesis (the data is noise) and hypotheses that some data is erroneous or entirely fictitious (to
counter misinformation). Multiple hypotheses may need to be combined to make a
prediction for an actual case  \citep{Poole:2012ab}. With an interconnected world, we don't need
to make conclusions from individual experiments, but can learn from
everyones experiences, which should enable more accurate hypotheses.

Maybe relational learning in its full generality is impossible, and  building separate applications for each domain is the best we
do.
The only way to determine this is to
try to build models that go across domains.

\section{Acknowledgements}
This work was supported by the Natural Sciences and Engineering Research Council of Canada (NSERC). Thanks to Sriraam  Natarajan for comments on an earlier version.
\bibliography{/Users/poole/World/bib/string,/Users/poole/World/bib/reason}
\clearpage
\appendix
\section{Appendix A: Proportion of Wikidata entities with few
  triples}\label{wikidata-app}
The aim of this appendix is to estimate the proportion of entities in Wikidata that are the
subject of few triples. I could not download all of
Wikidata, and it is not clear what a random sample is. Picking
entities at random results in nearly all of them being in very long tail of entities with
very  few triples, but this may not be representative of the
triples. Wikidata does not provide a tool to access a random triple.
To get a representative (if not random) sample, I selected the entities of the  Wikipedia featured
articles of March 2025
(\url{https://en.wikipedia.org/wiki/Wikipedia:Today%27s_featured_article/March_2025}) that had Wikidata pages. The entities were 
diverse including
Hughie Ferguson, a Scottish footballer in 1910s and 20s, 
Pulgasari, a North Korean film from 1985, the contemporary ballet Flight Pattern, Steele's Greenville expedition in the  American
Civil War, the sun, and a fictional character in the medical drama
series House. I tried a few other collections of entities and the results were
similar. This is not meant to be a statistically valid study.

For each entity (including reified entities), Wikidata
allows users to download triples
about that entity. For reified entities, the triples downloaded are the triples
of the subject of the reification. Define the local entities for 
entity $e$ to be those that appear as a subject in a triple for the
page for $e$ and for which Wikidata downloads the page for $e$.
For selected entity, I counted the
number of local entities that are the subject of fewer than 10 triples
and the number of entities that are the subject of 10 or more triples. See Figure
\ref{wikidata-stats-fig}. The number ranged from 5 for History of
infant schools in Great Britain and 6 for Your Girl (a 2005 song by Mariah Carey), 
to 520 for the composer and conductor Pierre Boulez, and 841 for the Sun. The entities
with the lowest number of triples
are just stubs; note that for Q133116210, the 11 triples included the name and 
labels in both English and Portuguese. The average number of local entities was
134 (or 108 if the top and bottom outliers were removed). 

\begin{figure}
  \begin{tabular}{lllll}
    Wikidata ID & \# triples & $<10$ & $\geq 10$ & name\\\hline
Q525 & 2605 & 841 & 2 & Sun \\
Q11813 & 1074 & 490 & 10 & James Madison \\
Q156193 & 859 & 520 & 5 & Pierre Boulez \\
Q485172 & 684 & 285 & 6 & Ann Arbor \\
Q192920 & 475 & 172 & 2 & Edward the Martyr \\
Q16267674 & 426 & 282 & 3 & Hotline Miami 2: Wrong Number \\
Q632511 & 325 & 113 & 4 & Seattle Sounders FC \\
Q556464 & 270 & 158 & 1 & Gertie the Dinosaur \\
Q1137583 & 242 & 121 & 1 & Pulgasari \\
Q822900 & 214 & 36 & 1 & geography of Ireland \\
Q4767002 & 210 & 89 & 1 & Anna Filosofova \\
 Q1349580 & 201 & 95 & 2 & Matthew Brettingham \\
Q5461994 & 201 & 112 & 1 & Flotilla \\
   Q3830799 & 161 & 63 & 1 & Leroy Chollet \\
  Q545151 & 141 & 51 & 2 & The Spy Who Loved Me \\
  Q3142353 & 125 & 50 & 5 & Hughie Ferguson \\
Q6529597 & 116 & 49 & 1 & Les Holden \\
Q931229 & 87 & 38 & 1 & USS Congress \\
Q3142456 & 78 & 41 & 1 & Hurricane Cindy \\
Q107132670 & 70 & 34 & 1 & Beverly White \\
Q3856283 & 70 & 30 & 1 & Michael Tritter \\
Q122330832 & 66 & 47 & 2 & All-American Bitch \\
Q830590 & 62 & 32 & 1 & Interstate 182 \\
Q130337417 & 32 & 17 & 1 & The True Record \\
Q3518834 & 58 & 20 & 1 & Territorial Force \\
Q98076985 & 29 & 21 & 1 & Flight Pattern \\
Q116245633 & 19 & 17 & 1 & Steele's Greenville expedition \\
Q133116210 & 11 & 5 & 1 & History of infant schools in Great Britain \\
Q131151772 & 9 & 6 & 0 & Your Girl \\
    \hline
    Average &308 &132 & 2 
  \end{tabular}
  \caption{For each top-level entity: the Wikidata ID, the number of triples with the entity as subject, the number of local
    entities that are the subject of fewer than 10 triples, the number with
    10 or more triples, and the (truncated)
    English name of the entity. Sorted by number of triples. These numbers were from 2025-08-19;
    they change over time as the knowledge base evolves. The Python code to
    generate the contents of this table is available from the author.}
  \label{wikidata-stats-fig}
\end{figure}
The
entities that are the subject of 10 or  more triples (other than the top-level entity)
are refied entities from relations with a large number of arguments; the maximum of
which is 17. About 1.5\% of the entities are the subject of 10 or more
triples. Fewer than 0.75\% of the entities are the subject of 18 or more
triples.

As the
number of triples for an entity increase, so does the number of
associated (local)
entities with few triples. As a knowledge graph gets bigger and
more filled out, the
proportion of entities with few triples increases. This goes against the
idea that we just need more data.

Wikidata did not include entities for all of the topics of the Wikipedia
features articles.
The March 12 Wikipedia article was for the 2020 season for Seattle
Sounders FC, which was not an entity in Wikidata, so we
used Seattle Sounders FC (Q632511). There was no Wikidata entry for
the March 14 page on the five-pound British gold coin, or the March 31
entry for  Apollo 15 postal covers incident. These were
omitted. There were 2 entities for ``geography of Ireland'', one for
the country and one for the island; the featured article was for the
island, so that is the entity used.

\section{Appendix B: An MRR agent}
The mean reciprocal rank (MRR) weights the $n$th result as
$\frac{1}{n}$.
As in discounting for MDPs and reinforcement learning, this can be
interpreted as the expected value for an agent with a probabilistic
stopping rule. An MRR agent always considers
the first item, considers the second item with probability
$\frac{1}{2}$, the third item with probability $\frac{1}{3}$, etc.
The MRR of an item corresponds to the probability that such an agent considers the
item.

The probability that an MRR agent considers exactly $n$ items is
\[\frac{1}{n}-\frac{1}{n+1} = \frac{1}{n(n+1)}.\]
The expected number of items this agent considers is
\[\sum_n n \frac{1}{n(n+1)} = \sum_n \frac{1}{n+1} = \infty\]
(This is the harmonic series, which is well known to not converge).
The problem is that the trillionth (and greater) element needs
to be reached  with a probability such that computing these large items
dominates the time. With finite
datasets, the divergence is just of theoretical interest. The agent needs to sometimes consider all
of the elements, but sometimes stops just short of this. It is
possible that such an agent models a real agent, but I have never seen any
evidence that this model fits the behavior of any real
agents. 

Note that an agent that
considers the $n$th item with probability $1/n^k$ for $k>1$ does
converge (similar to the Riemann zeta function). But then $k$ is an
arbitrary parameter which the MRR supposedly avoids.

Companies that present lists of results and get feedback about which
item(s) was selected or how many items were presented, such as Google, will
have data on the distribution of the number of items considered. It would be
better to use this distribution for evaluation than MRR, but it
probably depends on what is being presented (and whether 
predictions can be tested to determine if they are correct or what is wanted).
\end{document}